\documentclass[10pt,twocolumn,letterpaper]{article}
\usepackage{color}

\usepackage{cvpr}
\usepackage{times}
\usepackage{epsfig}
\usepackage{graphicx}
\usepackage{amsmath}
\usepackage{amssymb}
\usepackage{mathtools}
\usepackage{booktabs}

\usepackage{amsmath,amsfonts,bm}

\def\1{\bm{1}}

\def\rve{{\mathbf{e}}}

\DeclareMathAlphabet{\mathsfit}{\encodingdefault}{\sfdefault}{m}{sl}
\SetMathAlphabet{\mathsfit}{bold}{\encodingdefault}{\sfdefault}{bx}{n}

\usepackage[pagebackref=true,breaklinks=true,letterpaper=true,colorlinks,bookmarks=false]{hyperref}

\cvprfinalcopy 

\def\cvprPaperID{1836} 

\ifcvprfinal\fi
\begin{document}

\title{Memory Aggregation Networks for \\ Efficient Interactive Video Object Segmentation}
\author{Jiaxu Miao$^{1,2}$ \quad Yunchao Wei$^2$ \quad Yi Yang$^{2}$\footnotemark[2] \\
$^1$Baidu Research \quad $^2$ ReLER, University of Technology Sydney\\
\small \texttt{jiaxu.miao@student.uts.edu.au, \{yunchao.wei, yi.yang\}@uts.edu.au}
}

\maketitle
\thispagestyle{empty}
\renewcommand{\thefootnote}{\fnsymbol{footnote}}
\footnotetext[2]{Corresponding author.}
\footnotetext[3]{Part of this work was done when Jiaxu Miao was an intern at Baidu Research.}
\renewcommand*{\thefootnote}{\arabic{footnote}}
\begin{abstract}
{Interactive video object segmentation (iVOS) aims at efficiently harvesting high-quality segmentation masks of the target object in a video with user interactions. Most previous state-of-the-arts tackle the iVOS with two independent networks for conducting user interaction and temporal propagation, respectively, leading to inefficiencies during the inference stage. In this work, we propose a unified framework, named Memory Aggregation Networks (MA-Net), to address the challenging iVOS in a more efficient way. Our MA-Net integrates the interaction and the propagation operations into a single network, which significantly promotes the efficiency of iVOS in the scheme of multi-round interactions. More importantly, we propose a simple yet effective memory aggregation mechanism to record the informative knowledge from the previous interaction rounds, improving the robustness in discovering challenging objects of interest greatly. We conduct extensive experiments on the validation set of DAVIS Challenge 2018 benchmark. In particular, our MA-Net achieves the J@60 score of 76.1\% without any bells and whistles, outperforming the state-of-the-arts with more than 2.7\%.}
\end{abstract}
\begin{figure}[t]
\includegraphics[width=0.9\linewidth]{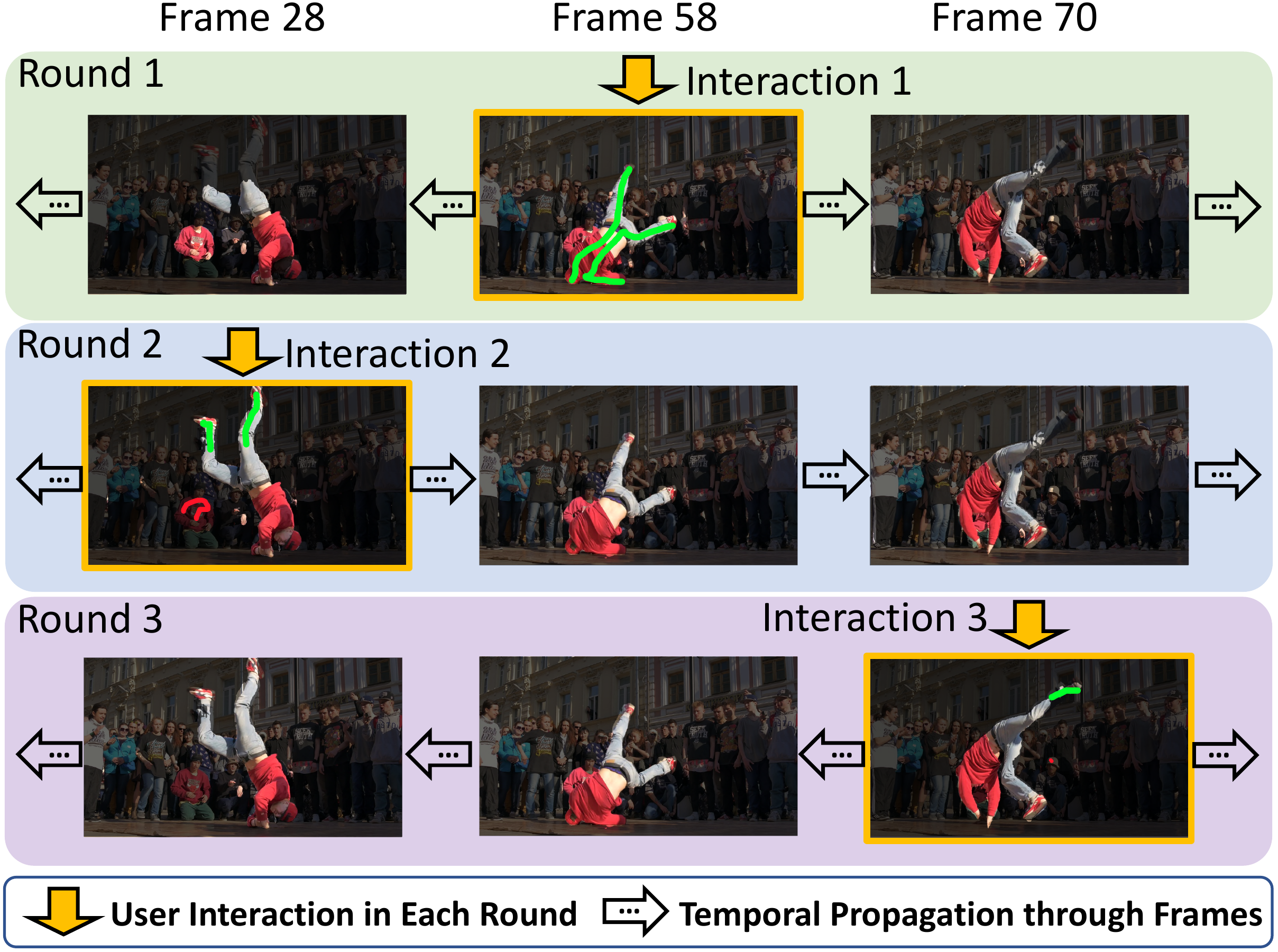}
\centering
\caption{Round-based iVOS. The mask of the target object is generated by user annotations at one frame (\eg green scribbles at frame 58), and the computed mask is propagated to generate the masks for the entire video. The user can refine the segmentation masks by repeatedly providing annotations on the false negative and false positive areas (\eg green and red scribbles at frame 28).}
\vspace{-3mm}
\label{fig:example_1}
\end{figure}
\section{Introduction}
Video object segmentation (VOS) aims at separating a foreground object from a video sequence and can benefit many important applications, including video editing, scene understanding, and self-driving cars. Most existing VOS approaches can be roughly divided into two settings: \emph{unsupervised} (no manual annotation) and \emph{semi-supervised} (give the annotation at the first frame). However, these two settings have their own limitations and are not realistic in practice: 1) unsupervised methods have no guiding signal for the user to select the object of interest, which is problematic especially for the multiple-object case; 2) semi-supervised methods need a fully annotated mask of the first frame, which is tedious to acquire (around 79 seconds per instance)~\cite{caelles20182018}. Furthermore, for both two schemes, users have no chance to correct those low-quality segments to meet their requirements.

Interactive video object segmentation (iVOS) overcomes the above-mentioned limitations by providing a user-friendly annotation form, \eg, scribbles. In this scheme, users can gradually refine the outputs by drawing scribbles on the falsely predicted regions. Previous iVOS methods~\cite{wang2005interactive,price2009livecut,bai2009video} utilize a rotoscoping procedure~\cite{bratt2012rotoscoping,li2016roto++}, where a user sequentially processes a video frame-by-frame. These methods are inefficient due to requiring a lot of user interactions at each frame.

Recently, Caelles~\etal~\cite{caelles20182018} propose a \emph{round-based interaction} scheme, as shown in Fig.~\ref{fig:example_1}. In this setting, users firstly draw scribbles on the target objects at one selected frame, and an algorithm is then employed to compute the segmentation masks for all video frames with temporal propagation. The procedures of user annotation and mask segmentation are repeated until acceptable results are obtained. Such a \emph{round-based interaction} scheme is more efficient since it requires fewer user annotations (only a few scribbles at one frame per round). Besides, it is flexible for users to control the quality of segmentation masks, since more rounds of user interactions will guarantee more accurate segmentation results. 

In this paper, we explore how to build an efficient interactive system to tackle the iVOS problem under the round-based interaction setting. 
While some recent deep learning based methods~\cite{oh2019fast,DAVIS2019-Interactive-2nd,DAVIS2018-Interactive-2nd,benard2017interactive,caelles20182018} have been proposed to deal with the round-based iVOS, there are several limitations: 1) the user interaction and the temporal propagation are usually processed by two independent networks~\cite{DAVIS2019-Interactive-2nd,benard2017interactive}; 2) the whole neural network has to start a new feed-forward computation in each interaction round~\cite{oh2019fast}, or needs post-processing~\cite{DAVIS2018-Interactive-2nd} to make a further refinement, which is time-consuming; 3) only the outputs of latest round are utilized to refine the segmentation results, while the informative multi-round interactions are usually ignored~\cite{DAVIS2019-Interactive-2nd}. 

Considering these limitations, we propose a \textbf{unified}, \textbf{efficient}, and \textbf{accurate} framework named Memory Aggregation Networks (MA-Net) to deal with the iVOS in a more elegant and effective manner. 
Concretely, our MA-Net integrates the interaction network and propagation network into a unified pixel embedding learning framework by sharing the same backbone. In this way, after extracting the pixel embedding with the shared backbone, the MA-Net adopts two ``shallow" convolutional segmentation heads to predict the object segments of the scribble-labeled frame and all the other frames, respectively. Under the round-based iVOS scheme, we only need to extract the pixel embedding of all the frames in the first round. In all the following rounds, these extracted embedding can be simply applied to make a further refinement with two ``shallow" segmentation heads, resulting in our MA-Net much faster than previous methods.
More importantly, we propose a simple yet effective memory aggregation mechanism, which is used to record informative knowledge of the user's interactions and the predicted masks during the previous interaction rounds. Such aggregated information makes the MA-Net robust to the target instances with a wide variety of appearances, improving the accuracy of our model greatly.


Our MA-Net is quantitatively evaluated on the interactive benchmark at the DAVIS Challenge 2018~\cite{caelles20182018}. On the DAVIS validation set, our MA-Net achieves the J@60 score of 76.1\% without any bells and whistles, such as introducing additional optical flow information~\cite{DAVIS2019-Interactive-2nd} or applying time-consuming CRF for post-processing~\cite{DAVIS2018-Interactive-2nd,krahenbuhl2013parameter}. In addition, our MA-Net can accomplish 7-round interactions within 60 seconds, which is more efficient than the state of the art one~\cite{oh2019fast} of 5-round interactions within 60 seconds.






\begin{figure*}[t]
\includegraphics[width=0.9\linewidth]{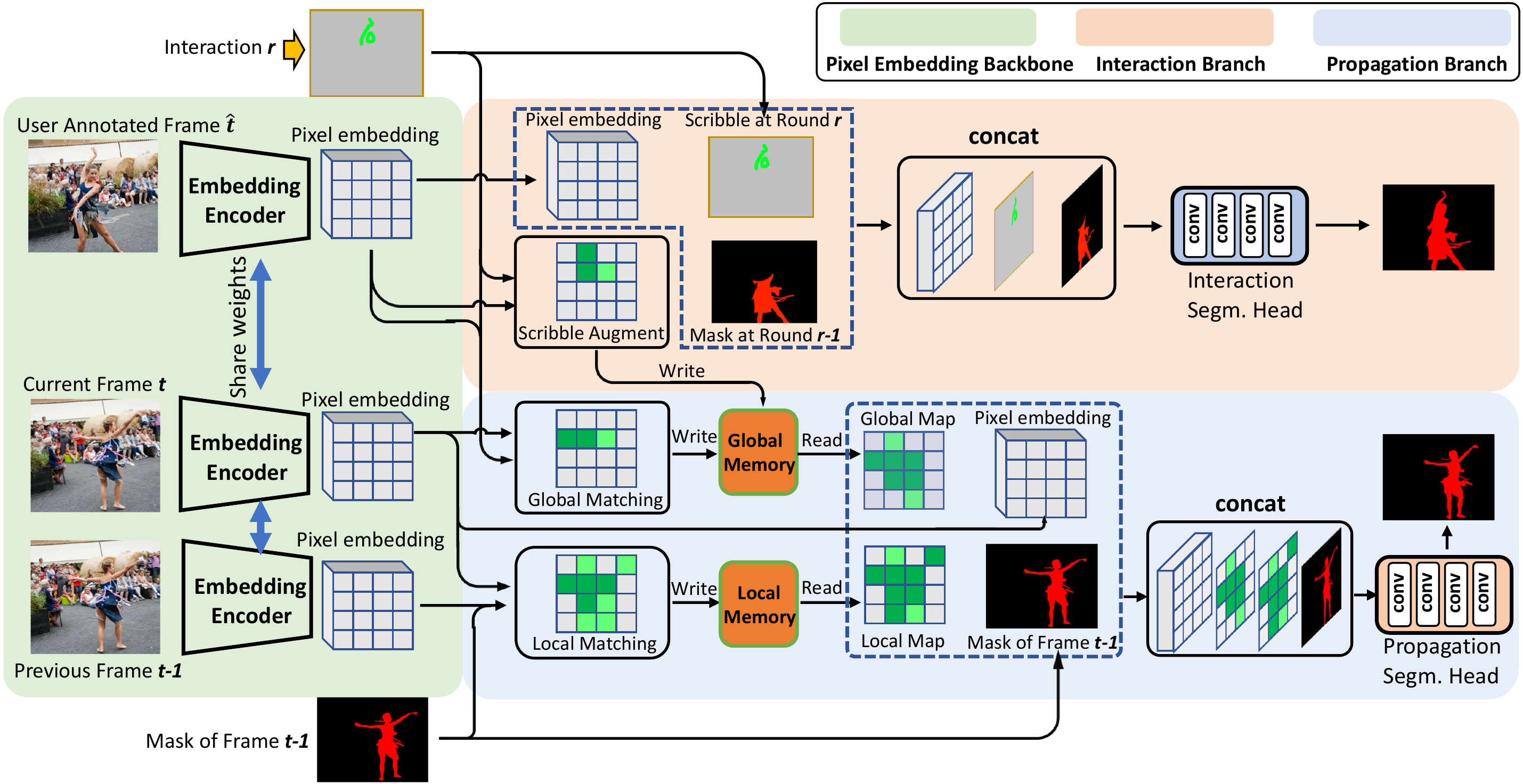}
\centering
\caption{The pipeline of our MA-Net, including the \textbf{pixel embedding backbone}, the \textbf{interaction branch}, and the \textbf{propagation branch}. During inference, the pixel embedding of all frames is extracted only once in the first round. The interaction branch employs ``shallow" convolutional layers to predict the mask of the interactive frame. The propagation branch uses a memory aggregation mechanism to record informative knowledge and ``shallow" convolutional layers to generate masks of other frames. In the matching processes shown in the figure, the deeper the green, the higher the probability of being predicted as the target object. Best viewed in color.}
\vspace{-3mm}
\label{fig:pipeline}
\end{figure*}
\section{Related Work}
\noindent \textbf{Unsupervised Video Object Segmentation.}
Unsupervised VOS does not need any user annotations. Most unsupervised segmentation models~\cite{ventura2019rvos,wang2019learning} learn to automatically segment visually salient objects based on the motion information or the appearance information. 
The limitation of unsupervised VOS is that users cannot select the object of interest.

\noindent\textbf{Semi-supervised Video Object Segmentation.}
Semi-supervised VOS employs the full annotation of the first frame to select the objects of interest. Many semi-supervised VOS methods~\cite{chen2018blazingly,hu2018videomatch,voigtlaender2019feelvos,wug2018fast,oh2019video,xu2019mhp,caelles2017one,voigtlaender2017online,maninis2018video,yang2018efficient,yang2020collaborative} have been proposed and achieve good performance. 

Some semi-supervised VOS approaches~\cite{caelles2017one,voigtlaender2017online,li2018video,luiten2018premvos} rely on fine-tuning using the first frame annotation at test time. For instance, OSVOS~\cite{caelles2017one} employs a convolutional neural network pre-trained for foreground-background segmentation and fine-tunes the model using first-frame ground truth when testing. OnAVOS\cite{voigtlaender2017online} and OSVOS-S~\cite{maninis2018video} further improve OSVOS by updating the network online using instance-level semantic information. 
PReMVOS~\cite{luiten2018premvos} integrates different networks with fine-tuning and merging, which achieves superior performance. Online fine-tuning methods achieve good performance, but poor efficiency due to the fine-tuning process at test time.

Recently, some VOS approaches without first-frame fine-tuning have been proposed and achieve very high speed and effectively. One type of these methods is \emph{propagation-based}~\cite{wug2018fast,yang2018efficient,bao2018cnn}, which usually takes as input the combination of the image and predicted segmentation mask of the previous frame.  For instance, RGMP~\cite{wug2018fast} employs a siamese architecture network. One stream encodes the feature of the target frame and the mask of the previous frame while another stream encodes the first frame together with its given ground truth. 
Another type of fine-tuning free methods is \emph{matching-based}~\cite{chen2018blazingly,hu2018videomatch,voigtlaender2019feelvos,wang2019ranet}, which utilizes the \textbf{pixel embedding learning}. For instance, PML~\cite{chen2018blazingly} learns a pixel embedding space by a triplet loss together with a nearest neighbor classifier. VideoMatch~\cite{hu2018videomatch} proposes a soft matching mechanism by calculating similarity score maps of matching features to generate smooth predictions. FEELVOS~\cite{voigtlaender2019feelvos} employs pixel-wise embedding together with a global and a local matching mechanism. By considering foreground-background integration, CFBI~\cite{yang2020collaborative} achieves the new state of the art.
Our method is inspired by FEELVOS~\cite{voigtlaender2019feelvos}, and utilizes the global and local matching maps to transfer information of the scribble-annotated and previous frame to the target frame. 


\noindent\textbf{Interactive Video Object Segmentation.}
In the interactive VOS setting, users can provide various types of inputs (\eg points, scribbles) to select the objects of interest and refine the segmentation results by providing more interactions. 
Previous interactive methods~\cite{wang2005interactive,price2009livecut,bai2009video}, either use hand-crafted features or need a lot of interactions, can not achieve a good performance or efficiency.

Recently, some round-based deep learning methods~\cite{oh2019fast,DAVIS2019-Interactive-2nd,DAVIS2018-Interactive-2nd,benard2017interactive} for iVOS have been proposed. Benard~\etal~\cite{benard2017interactive} and Heo~\cite{DAVIS2019-Interactive-2nd} treat the interactive VOS as two sub-tasks: using the scribbles to generate segmentation masks, and using the generated mask to infer masks of other frames as semi-supervised VOS. Oh~\cite{oh2019fast} uses two networks, interaction and propagation, to tackle these two sub-tasks. These two networks are connected both internally and externally. These methods~\cite{benard2017interactive,DAVIS2019-Interactive-2nd,oh2019fast} have several limitations: (1) they use two independent networks without shared weights, and need new feed-forward computation in each interaction round~\cite{oh2019fast,DAVIS2019-Interactive-2nd}, making it inefficient when rounds grow up; (2) they do not utilize the multi-round information adequately. Recently, Oh~\cite{oh2019video} proposes a space-time memory mechanism to store informative knowledge and achieves state-of-the-art performance. Different from our memory mechanism, they need complicated key-value computation. Besides, they also need new feed-forward computation in each interaction round, which is time-consuming.

\section{Method}
Round-based iVOS aims at cutting out the target objects in all frames of a video given user annotations (\eg scribbles) on one frame. Users can provide additional feedback annotations on a frame after reviewing the segmentation results to refine the segmentation mask of the next round.   Previous methods~\cite{DAVIS2019-Interactive-2nd,oh2019fast,benard2017interactive} chose to adopt two independent neural networks (interaction and propagation) without shared weights or connect two networks by medial layers, which usually affects the inference efficiency. In this paper, we deal with the two sub-tasks (interaction and propagation) under a unified pixel embedding learning framework. 

To this end, we propose MA-Net, which contains three modules: a pixel embedding encoder, an interaction branch, and a propagation branch, as shown in Fig.~\ref{fig:pipeline}.  The pixel embedding encoder takes the RGB frames of the given video as inputs and encodes each pixel into an embedding vector. The interaction branch leverages the user's annotations (scribbles) and the pixel embedding of the user-annotated frame to generate the instance segmentation mask. The propagation branch propagates the informative knowledge of the user-annotated frame and the previous frame to the current frame using the pixel embedding. Both the two branches share weights of the pixel embedding encoder, and then employ two ``shallow" networks with several convolutional layers as the segmentation heads, respectively. The pixel embeddings of all frames are extracted only in the first interaction round. During the refinement process in the following rounds, only the two ``shallow" segmentation heads are used, making our MA-Net more efficient than previous methods. 
In this paper, we denote the current processing frame as the $t^{th}$ frame, the previous frame as the $(t-1)^{th}$ frame, and the user-annotated frame as the $\hat{t}^{th}$ frame. Pixels of the current processing frame are denoted as $p$, and pixels annotated or predicted to belong to the target object $o$ as $q$.
In the following, we will describe each of the modules in more detail.
\begin{figure}[t]
\includegraphics[width=0.85\linewidth]{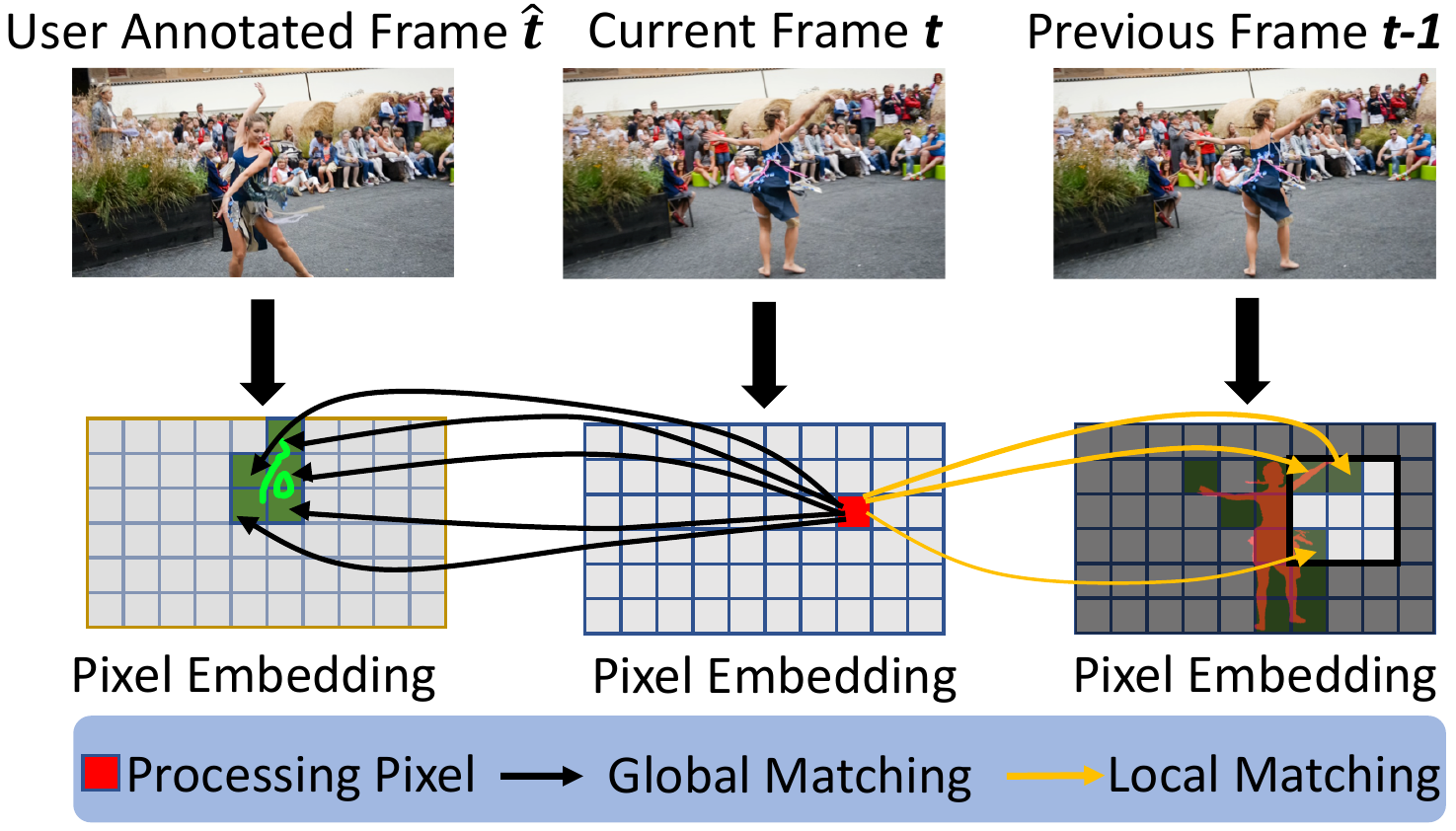}
\centering
\caption{Global matching and local matching process. For each pixel in the current processing frame at time $t$, distances are calculated with pixels of the target object annotated by scribbles (global map) or predicted mask (local map) and the smallest value of distances (nearest neighbor) are used to construct the matching map.}
\vspace{-3mm}
\label{fig:match}
\end{figure}

\textbf{Pixel Embedding Encoder.}
The purpose of pixel embedding learning is to learn an embedding space where pixels belonging to the same object are close while pixels belonging to different objects are far away. 
 We employ the DeepLabv3+ architecture~\cite{chen2018encoder} based on ResNet101~\cite{he2016deep} as our backbone, and add an embedding layer consisting of one depth-wise separable convolution with a kernel size of $3 \times 3$. The stride of the pixel embedding feature is 4, and the dimension is 100.
For each pixel $p$ in the input RGB frame, we learn a semantic embedding vector $\rve_p$ in the learned embedding space.  In this paper, we encode the pixel embedding into a Euclidean space, where the Euclidean norm between two pixels in the same object is expected to be small. Similar to ~\cite{fathi2017semantic,voigtlaender2019feelvos}, we define the distance between pixels $p$ and $q$ in terms of their corresponding embedding vectors $\rve_p$ and $\rve_q$ as
\vspace{-3mm}
\begin{equation}
    d(p,q)=1-\frac{2}{1+\exp({\left\|{\rve_p-\rve_q}\right\|}_2^2)}.
\end{equation}
This operation aims at normalizing the pixel distance between 0 and 1. 
We follow the strategy of FEELVOS~\cite{voigtlaender2019feelvos} to employ the pixel distances as a soft cue, which is further refined by two ``shallow" segmentation heads. 
\begin{figure}[t]
\includegraphics[width=0.8\linewidth]{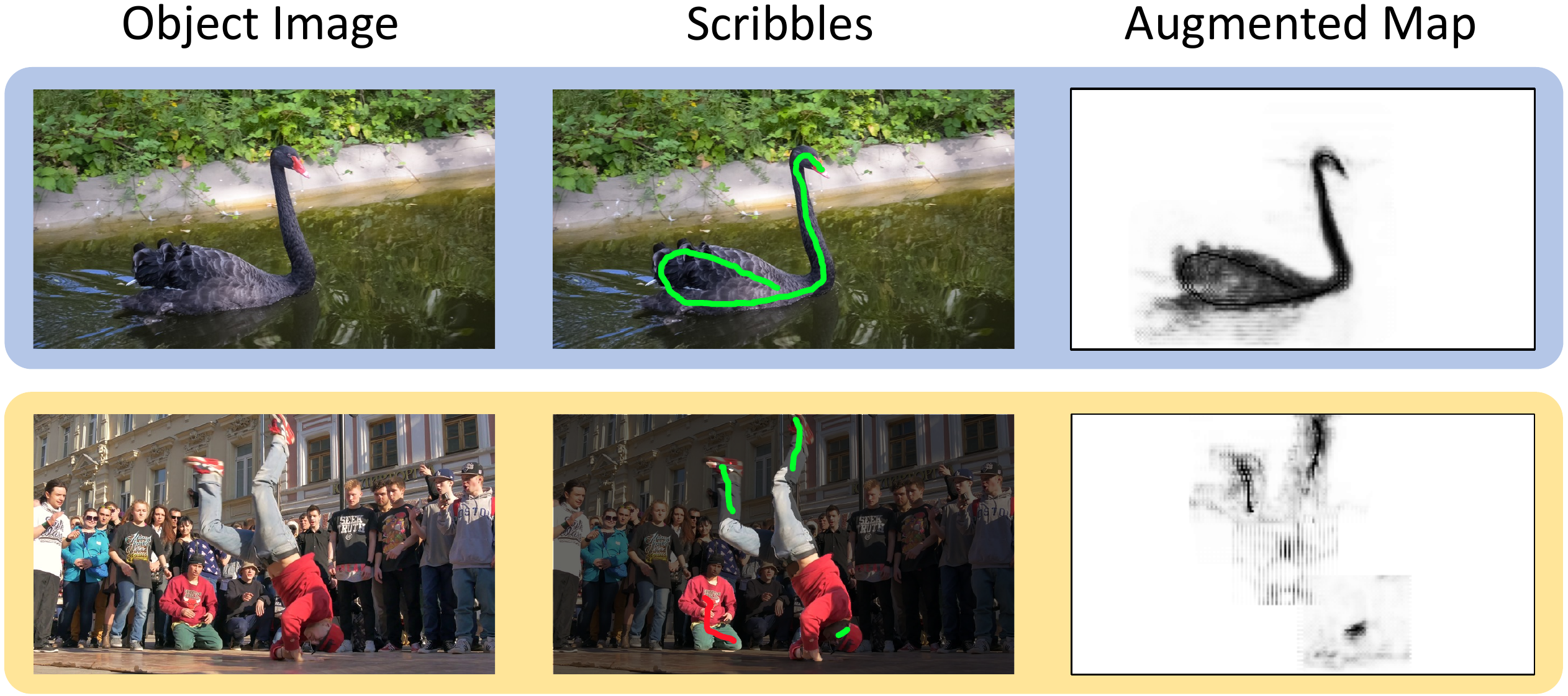}
\centering
\caption{Examples of the augmented map computed by the pixel embedding and scribbles.}
\vspace{-1.5em}
\label{fig:scribbleaug}
\end{figure}

\begin{figure*}[t]
\includegraphics[width=0.45\linewidth,height=5cm]{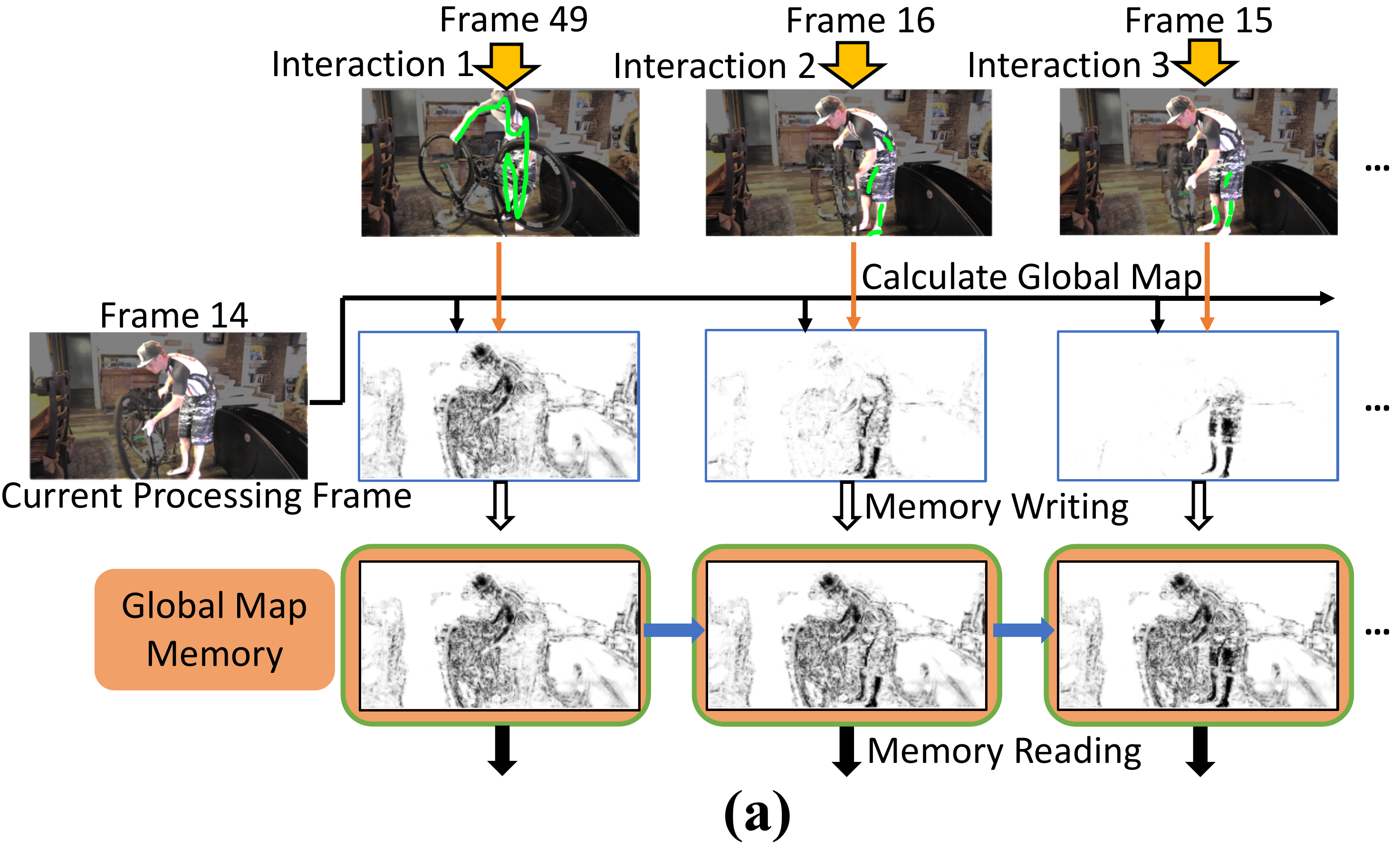}
\includegraphics[width=0.45\linewidth]{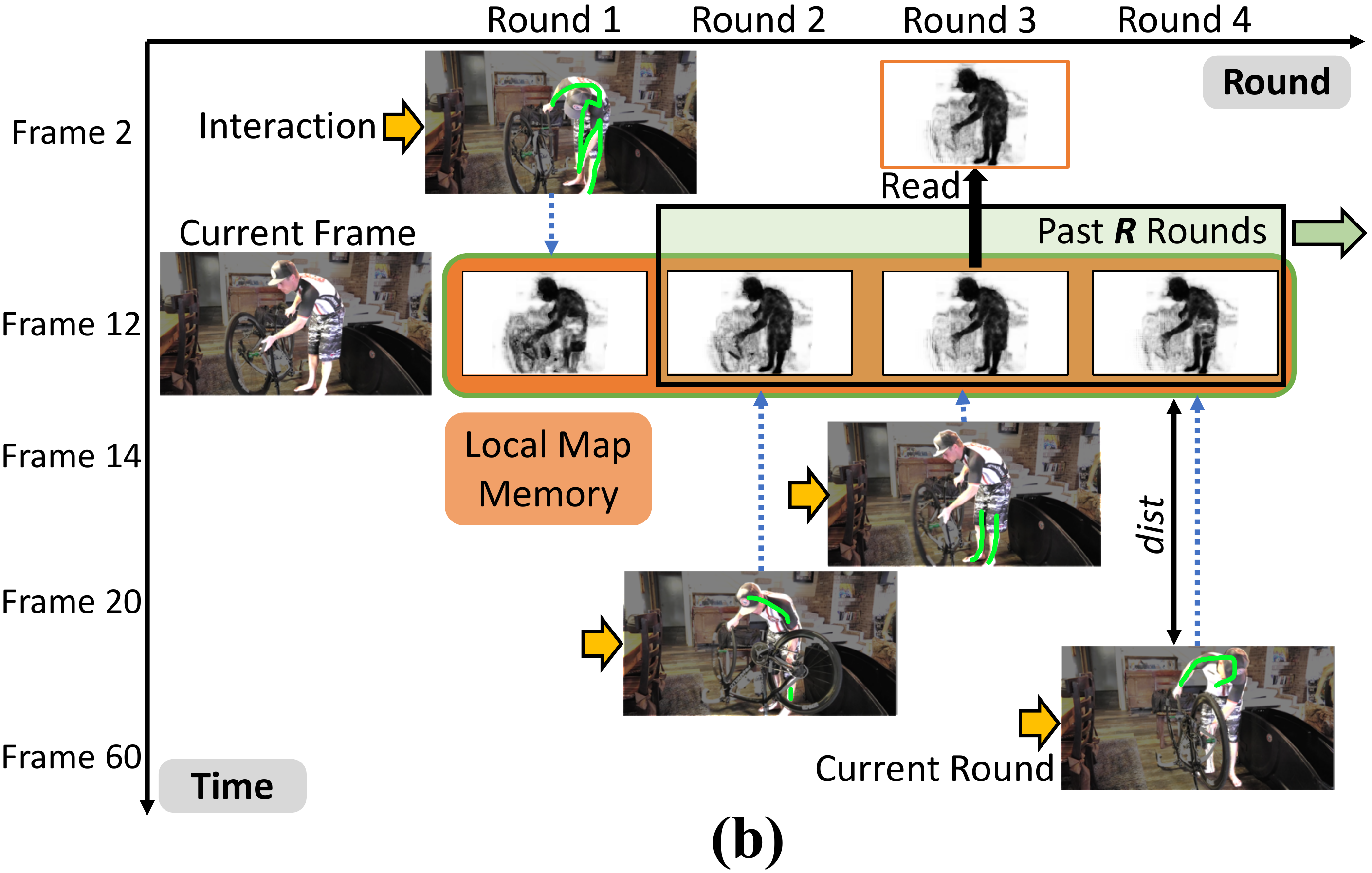}
\centering
\caption{\textbf{(a)} The \textbf{global map memory mechanism}. The global map in the propagation branch and the augmented map in the interaction branch are recorded and aggregated in the memory. \textbf{(b)} The \textbf{local map memory and forgetting mechanism}. The local map in each interaction round is recorded in the memory, and the nearest map of time in past $R$ rounds is read to compute the masks. Local maps of early interaction rounds are forgotten with rounds growing up. The blue arrows denote the temporal propagation.}
\vspace{-1.5em}
\label{fig:memory}
\end{figure*}

\textbf{Propagation Branch.}
The propagation branch aims at propagating information from the user-annotated frame and the previous frame to predict the segmentation mask of the target object at the current frame. Following FEELVOS~\cite{voigtlaender2019feelvos}, we employ the global and local matching map as the soft cues of the user-annotated frame and the previous frame, respectively. The matching processes of the global and local maps are shown in Fig.~\ref{fig:match}. Different from FEELVOS~\cite{voigtlaender2019feelvos}, our MA-Net proposes to employ a memory aggregation mechanism to record and aggregate the informative knowledge during the previous multiple interaction rounds, which is specially designed for iVOS. 

\emph{Global Map Memory.}
Let $\mathbb{P}_{t}$ denotes the set of all pixels of the current $t^{th}$ frame and  
$\mathbb{P}_{\hat{t},o,r}$ denotes the set of user-annotated pixels of the interactive $\hat{t}^{th}$ frame in the $r^{th}$ round of interaction. As show in the left of Fig.~\ref{fig:match}, for each pixel $p \in \mathbb{P}_{t}$, we can calculate the distance of its nearest neighbour in $\mathbb{P}_{\hat{t},o,r}$ to construct a global matching distance map, which is defined by
\begin{equation}
    \textbf{G}_{t,r}(p)=\min_{q \in \mathbb{P}_{\hat{t},o,r}}d(p,q).
\end{equation}

Different from the semi-supervised VOS who obtains a fully annotated frame, the interactive setting only provides a small number of scribble annotations to the objects of interest in each round. 
Therefore, the produced global matching map in one round is usually insufficient to discover the entire target object. To tackle this problem, we build a global memory unit to record and aggregate the historical global matching maps to enrich the information of the target object. Let $\textbf{M}^g \in \mathbb{R}^{n,o,h,w}$ denotes the global map memory, where $n,o,h,w$ denotes the total number of video frames, the target object, the height and width of the embedding feature maps, respectively. Consider that the range of the values in the matching map is from $0$ to $1$, where the value of pixels closer to $0$ is more likely to belong to the selected object and vice versa. We initialize $\textbf{M}^g$ with $1$ and update $\textbf{M}^g$ by preserving the minimum value of each pixel in different interaction rounds. We demonstrate the updating process of the global map memory in Figure.~\ref{fig:memory} (a). Formally, for the round of $r$ and the frame at time $t$, $\textbf{M}^g$ is \emph{written} by
\begin{equation}
    \textbf{M}^g_{t,r}=\min(\textbf{M}^g_{t,r-1},\textbf{G}_{t,r}).
\end{equation}
When we \emph{read} the accumulated global map of round $r$, we directly use the updated global map memory $\textbf{M}^g_{t,r}$.

\emph{Local Map Memory and Forgetting.}
Since the motion between two adjacent frames is usually small, to take advantage of the information of predicted mask from the previous frame, we further introduce the local matching map~\cite{voigtlaender2019feelvos}.
To avoid false-positive matches as well as save computation time, we only calculate the matching distance map with a small local region. Let $\mathbb{P}_{t-1,o}$ denote the pixels of frame at time $t-1$ which are predicted to be the object $o$. $\mathbb{N}(p)$ denotes the neighborhood set of pixel $p$, which contains pixels at most $k$ pixels far away from $p$. As shown in the right of Fig.~\ref{fig:match}, for each pixel $p$ belonging to the frame at time $t$, we can then compute the local matching distance map $\textbf{L}_{t,r}$ by
\begin{equation}
        \textbf{L}_{t,r}(p)=\begin{cases}
        \min_{q \in \mathbb{P}_{t-1,o}^N}d(p,q) \quad \text{if  }   \mathbb{P}_{t-1,o}^N \neq \varnothing \\
    1 \qquad \qquad \qquad \quad \text{   otherwise,} \\
    \end{cases}
\end{equation}
where $\mathbb{P}_{t-1,o}^N \coloneqq \mathbb{P}_{t-1,o} \cap\mathbb{N}(p)$ is the intersection set of the previous frame pixel set $\mathbb{P}_{t-1,o}$ and the neighbour set $\mathbb{N}(p)$.

Different from the scribble annotations provided by users, the mask information of the previous frame is unreliable since the segmentation mask of the previous frame is predicted by the algorithm. In practice, we found that the error will accumulate due to drifts and occlusions during the propagation. The segmentation result will get worse if the current frame \emph{far away} from the user-annotated frame. Therefore, to prevent the error accumulation, we additionally build a local memory unit $\textbf{M}^l \in \mathbb{R}^{n,r,o,h,w}$ to record the historical local matching maps in the previous interaction rounds. Formally, the local map $\textbf{L}_{t,r}$ for the $t^{th}$ frame in round $r$ is \emph{written} into the local memory by
\begin{equation}
    \textbf{M}^l_{t,r}= \textbf{L}_{t,r},
\end{equation}
which means that the writing process of the local memory is simply recording.

When reading from the local memory, for the current $t^{th}$ frame, we calculate the distance of time to the user-annotated $\hat{t}^{th}$ frame of each round $r$, $dist_r=|t-\hat{t}_r|$, and select the nearest one to the user-annotated frame as the final local map.
However, with the interactive round grows up, the accuracy of segmentation becomes better and better. For instance, a processing frame using the local map of round $8$, although far away from the user-annotated frame in this round, may be better than using the local map of round $1$ adjacent to the user-annotated frame. Hence we employ a \emph{forgetting mechanism} by using the nearest local map to the user-annotated frame in only past $R$ rounds. Local maps of early interaction rounds will be forgotten. $R=1$ means we only use local maps of the current round and do not employ the memory mechanism. The local map memory and forgetting mechanism is shown in Fig.~\ref{fig:memory} (b). Formally, denote the final local map for the current $t^{th}$ frame in round $r^*$ as $\textbf{L}'_{t,r^*}$, then $\textbf{L}'_{t,r^*}$ is \emph{read} from $\textbf{M}^l$ by
\begin{equation}
    \textbf{L}'_{t,r^*} = \textbf{M}^l_{t,r'}, r' = \arg\min_r |t-\hat{t}_r|\ and\ |r'-r^*| \leq R
\end{equation}

We utilize the propagation head with four convolutional layers to predict a one-dimensional map of logits for each selected object. The propagation head takes as input the concatenation of the pixel embedding, the global and local matching map read from memories, and the predicted mask of the previous frame. We stack the logits, apply softmax over the object dimension to obtain the probability map for each pixel.

\textbf{ Interaction Branch.}
The interaction branch aims at generating a segmentation mask of the user-annotated frame (interactive frame) given user annotations. As shown in Fig.~\ref{fig:pipeline}, for generating the segmentation mask of the interactive frame in the current round, we concatenate the pixel embedding, the scribbles and the predicted mask from last round along the channel dimension, and use an interaction segmentation head with four convolutional layers to generate the segmentation logits of the target object $o$. For the multi-object cases, the interaction segmentation head extracts one-dimensional feature maps of logits for all objects, which are then stacked together to obtain the probability map for each pixel by applying the softmax operation over the object dimension.

In iVOS, the interaction branch need not only generate the segmentation mask of the interactive frame in the current round but also record and accumulate informative knowledge of the scribbles for improving the segment results of this frame in the next rounds.
We propose a matching map to \emph{augment} the incomplete scribbles by mining the property of the pixel embedding space, and record the augment map into the global memory $\textbf{M}^g$. 
In the pixel embedding space, the pixels close to the annotated pixels have a higher probability of belonging to the same object.
Similar to the local map proposed in the propagation branch, we employ a matching distance map to augment the scribbles. Suppose $\mathbb{P}_{\hat{t}}$ denote the set of all pixels (with a stride of 4 in the embedding space) of the user-annotated frame at time $\hat{t}$ and $\mathbb{P}_{\hat{t},o}$ denote the set of scribble-annotated pixels belonging to the target object $o$.
For each pixel $p \in \mathbb{P}_{\hat{t}}$, we compute the distance of its nearest neighbor in the annotated pixels $\mathbb{P}_{\hat{t},o}$ to construct the matching distance map.
To avoid introducing the unexpected noisy pixels that are similar to the annotated ones but with large spatial distances, for each pixel $p \in \mathbb{P}_{\hat{t}}$, we only consider those pixels within its local neighborhood as the searching candidates. 
We denote $\mathbb{N}(p)$ as the neighbourhood set of $p$, where $\mathbb{N}(p)$ contains pixels at most $k$ pixels far away from $p$. 
Therefore, the augmented map $\textbf{A}_{\hat{t}}(p)$ for pixel $p$ is defined by

\vspace{-4mm}
\begin{equation}
    \textbf{A}_{\hat{t}}(p)=\begin{cases}
        min_{q \in \mathbb{P}_{\hat{t},o}^N}d(p,q) \quad \text{if  }   \mathbb{P}_{\hat{t},o}^{N} \neq \varnothing \\
    1 \qquad \qquad \qquad \quad \text{   otherwise,} \\
    \end{cases}
\end{equation}
where $\mathbb{P}_{\hat{t},o}^N \coloneqq \mathbb{P}_{\hat{t},o}\cap\mathbb{N}(p)$ is the intersection set of the scribble-annotated set $\mathbb{P}_{\hat{t},o}$ and the neighbourhood set $\mathbb{N}(p)$.
Fig.~\ref{fig:scribbleaug} shows the comparison of the scribbles and the augmented maps, and we can find that the augmented map contains more information about the selected objects. The augmented map $\textbf{A}_{\hat{t}}$ will be recorded and aggregated in the global map $\textbf{M}^g$. For the interactive frame at the time $\hat{t}$ in the round of $r$, $\textbf{M}^g$ is updated by
\begin{equation}
    \textbf{M}^g_{\hat{t},r}=\min(\textbf{M}^g_{\hat{t},r-1},\textbf{A}_{\hat{t},r}).
\end{equation}
This operation can benefit the segmentation result of the interactive frame in next rounds.

\begin{figure}[t]
\includegraphics[width=0.95\linewidth]{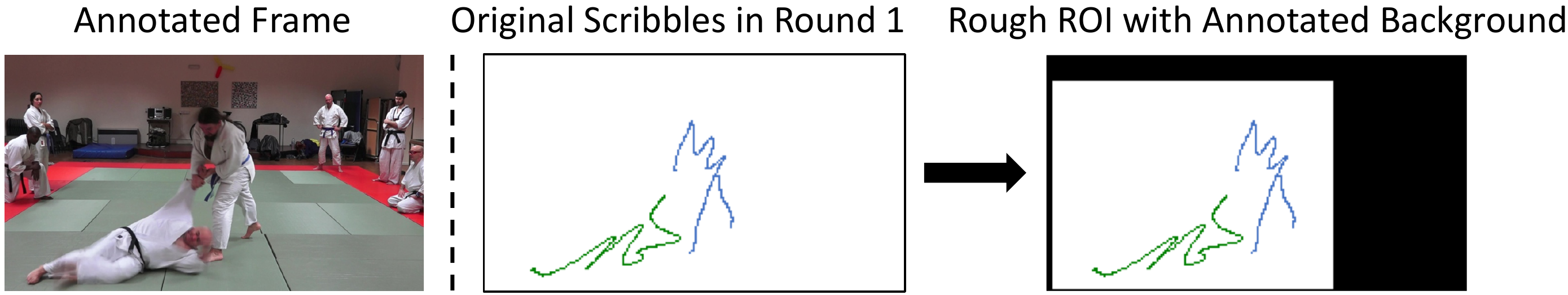}
\centering
\caption{In the first round, there are no annotations of the background. We use a rough ROI and annotate pixels out of ROI as the background (black area). Green and blue scribbles annotate the first and second objects, respectively.}
\vspace{-1em}
\label{fig:roughROI}
\end{figure}

\section{Experiments}
\subsection{Training and Inference}
\textbf{Training Procedure.}
We employ a two-stage training procedure to train our MA-Net. In Stage 1, we train the propagation branch with the pixel embedding encoder. To simulate the video propagation process, we randomly select three frames from one training video as a training batch. One of the frames serves as the reference frame, \ie, it plays the role of the frame annotated by scribbles. Two adjacent frames serve as the previous frame and the current processing frame.  Some methods~\cite{oh2019fast,DAVIS2019-Interactive-2nd} leverage the synthesized scribbles for the reference frame to train the propagation network. However, the synthesized scribbles are all densely generated from the ground-truth masks. After performing the training for a large number of iterations, ground-truth masks are actually used. Since the propagation branch is trained independently and densely generating synthesized scribbles from groundruth in an online manner is often time-consuming, we directly use the ground-truth instance mask of the reference frame. 
In practice, we found that the reference frame using ground truth achieves similar performance to using the synthesized scribbles during training.

In Stage 2, after training the pixel embedding encoder and the propagation branch, we fixed the pixel embedding encoder and trained the interaction branch. It is not feasible to collect a large number of scribbles annotated by users. Therefore, we train our model with synthesized scribbles. In the first round, we use the scribbles of the training set provided by the DAVIS Challenge 2018~\cite{caelles20182018}. In the following rounds, scribbles are synthesized within false negative and false positive areas. There is a gap between the first round and the following rounds since the first round only provides positive scribbles while following rounds provide both positive and negative scribbles. Hence we use the background label as the mask of the previous round for the first round.

\textbf{Inference.}
We follow the round-based interactive setting of the DAVIS Challenge 2018. In the first round, users provide positive scribbles and no negative scribbles. 
To eliminate the gap between training and testing, we use a rough Region of Interest (ROI) that contains all positive scribbles and enlarge ROI by enough space to make sure it contains all parts of the target object. Then we annotate all the pixels out of the enlarged ROI as the background (Fig.~\ref{fig:roughROI}). 
We extract the pixel embedding of each frame and utilize the interaction branch and propagation branch to generate segmentation masks of the target video. In the following round, users annotate the frame of the video with the worst performance using scribbles. Our model extracts the pixel embeddings of all frames for only once in the first round. The extracted pixel embeddings are further employed to compute the refined segmentation masks with the interaction and propagation heads in the following rounds, leading to our MA-Net more efficient than previous methods.
\begin{figure*}[t]
\includegraphics[width=0.8\linewidth]{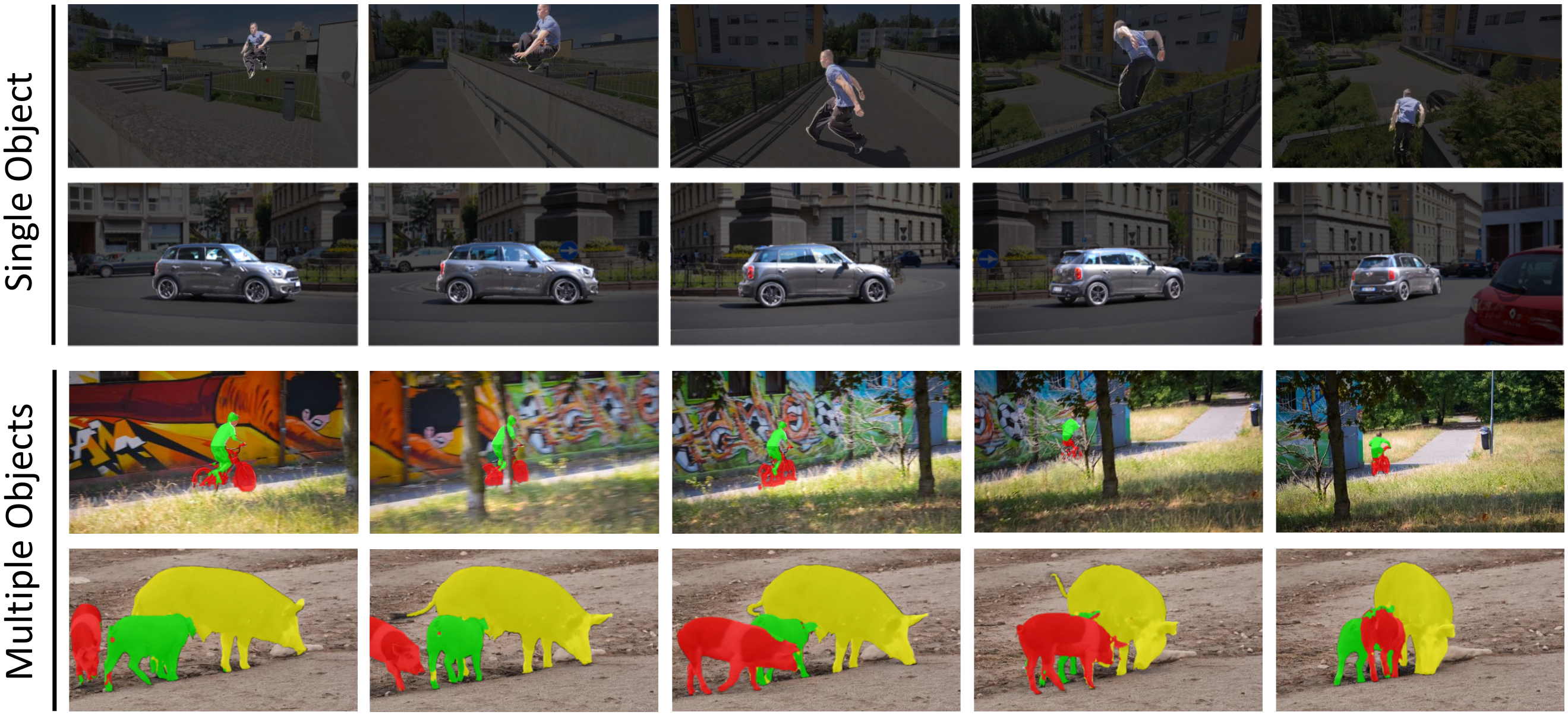}
\centering
\caption{The qualitative results on the DAVIS-2017 validation set. All the user interactions are automatically simulated by the robot agent provided by~\cite{caelles20182018}. All result masks are sampled after 8 rounds.}
\vspace{-3mm}
\label{fig:result_example}
\end{figure*}

\textbf{Implementation Details.} We use the DeepLabv3+ architecture~\cite{chen2018encoder} based on ResNet101~\cite{he2016deep} as our backbone, which produces an output feature maps with a stride of 4. On the top of the backbone, we add an embedding layer consisting of one depth-wise separable convolution with a kernel size of $3 \times 3$. The dimension of the pixel embedding is 100 advised by~\cite{voigtlaender2019feelvos}.

For the interaction and propagation segmentation heads, we employ four depth-wise separable convolutional layers with a dimension of 256, a kernel size of $7 \times 7$ for the depth-wise convolutions, a batch normalization operation and a ReLU activation function. Finally, a $1 \times 1$ convolution is employed to extract the predicted logits.

When computing the local matching map, we downsample the pixel embedding by a factor of 2 for computational efficiency. In practice, we set the local window size as $k=12$ in this paper, considering the trade-off between accuracy and efficiency. We utilize SGD optimization with a learning rate of 0.0007 and a batch size of 2. We employ the adaptive bootstrapped cross-entropy loss~\cite{pohlen2017full}, which takes into account 100\% to 15\% hardest pixels from step 0 to step 50000 for computing the loss. All input images are augmented by random flipping, scaling, and cropping. The input size is $416 \times 416$ pixels. When processing the training of the first stage, we initialize the weights of the backbone with the weights pre-trained on ImageNet~\cite{deng2009imagenet} and COCO~\cite{lin2014microsoft}, and we train the pixel embedding encoder and the propagation head on the training set of DAVIS~\cite{pont20172017} for 100000 steps. When training our model in the second stage, we use a round-based training with three rounds per circle. The first round uses only the positive scribbles while the following two rounds use both the positive and negative scribbles and the previous round masks. We train the second stage on the training set of DAVIS~\cite{pont20172017} for 80000 steps.

\subsection{Results}
Evaluating iVOS quantitatively is difficult since the user input is directly related to the segmentation results, and different users may provide different scribbles. To tackle this problem, Caelles~\etal~\cite{caelles20182018} proposes a robot agent service to simulate human interaction for a fair comparison.

\textbf{Quantitative Results.} 
To fairly compare our MA-Net with the state-of-the-art methods, we evaluated our model on the DAVIS validation set following the interactive track benchmark in the DAVIS Challenge 2018~\cite{caelles20182018}. In this benchmark, a robot agent interacts with each model for 8 rounds, and the model is expected to compute masks within 30 seconds per interaction for each object. There are two evaluation metrics: area under the curve (AUC) and Jaccard at 60 seconds (J@60s). AUC is designed to measure the overall accuracy of the evaluation. J@60 measures the accuracy with a limited time budget (60 seconds). Table.~\ref{table:compare} shows the comparison of our method and previous state-of-the-art iVOS methods. 
Comparing with the best competing method Heo~\cite{DAVIS2019-Interactive-2nd}, according to accuracy, our method surpasses it by +4.7\% AUC.   Comparing with the best competing method Oh~\etal~\cite{oh2019fast}, according to efficiency, our method surpasses it by +2.7\% J@60s.  Besides, our model does not use any bells and whistles such as optical flow, post-processing (CRF), or additional video training set, \ie, YoutubeVOS~\cite{xu2018youtube}. In addition, our MA-Net can accomplish 7-round interactions within 60seconds, which is more efficient than the state of the art one~\cite{oh2019fast}  of  5-round interactions within 60  seconds~\footnote{To fairly compare the efficiency, we test our model on a 1080Ti GPU following Oh~\cite{oh2019fast}}. In summary, our MA-Net outperforms previous methods in both accuracy and efficiency.

\begin{table}[t]
\centering
\small
\setlength{\tabcolsep}{4.5pt}
\begin{tabular}{c|c c c|c c}
\hline\hline
Method & +OF & +CRF & +YV & AUC &J@60\\ 
\hline
Najafi~\etal~\cite{DAVIS2018-Interactive-2nd} & & \checkmark &  & 0.702 &0.548 \\
Heo~\etal~\cite{DAVIS2019-Interactive-2nd} & & & \checkmark & 0.698 & 0.691 \\
Heo~\etal~\cite{DAVIS2019-Interactive-2nd} & \checkmark & & \checkmark & 0.704 & 0.725 \\
Oh~\etal~\cite{oh2019fast} & & & \checkmark & 0.691 & 0.734 \\
\hline
\textbf{MA-Net(Ours)} & & & & \textbf{0.749} & \textbf{0.761} \\
\hline
\end{tabular}
\caption{Comparison of our MA-Net with the previous methods on the validation set in DAVIS2017. The entries are ordered according to the J@60 score. +OF denotes using optical flow, +CRF denotes using the CRF~\cite{krahenbuhl2013parameter} as post-processing and +YV denotes using additional YoutubeVOS training set~\cite{xu2018youtube} when training.}
\vspace{-1.5em}
\label{table:compare}
\end{table}

\begin{table}[t]
\centering
\small
\setlength{\tabcolsep}{4.9pt}
\begin{tabular}{c|c c c c}
\hline\hline
Local window size $k$  & 6 &9 &12 &15\\ 
\hline
AUC & 0.724 & 0.737 & 0.749 & 0.748 \\ 
\hline
J@60 &0.730 & 0.753 & 0.761 & 0.761 \\
\hline 
\end{tabular}
\caption{The impact of the local window size $k$.}
\vspace{-1em}
\label{table:localk}
\end{table}
\textbf{Qualitative Results.} 
Fig.~\ref{fig:result_example} shows qualitative results on the DAVIS 2017 validation set.  It can be seen that our MA-Net produces accurate segmentation masks in multiple cases of large variance, including the single object condition and multiple objects condition. Qualitative results also show that our method can handle the occlusion issue (the $3$rd row). In some difficult cases, \eg, a video contains multiple objects of the same class and the objects are occluded by each other (pigs in the $4$th row), our method may make mistakes in some similar parts of different objects. This is most likely because the pixel embedding vectors of similar parts are close to each other.

\subsection{Ablation Study}
\begin{figure}[t]
\includegraphics[width=0.85\linewidth]{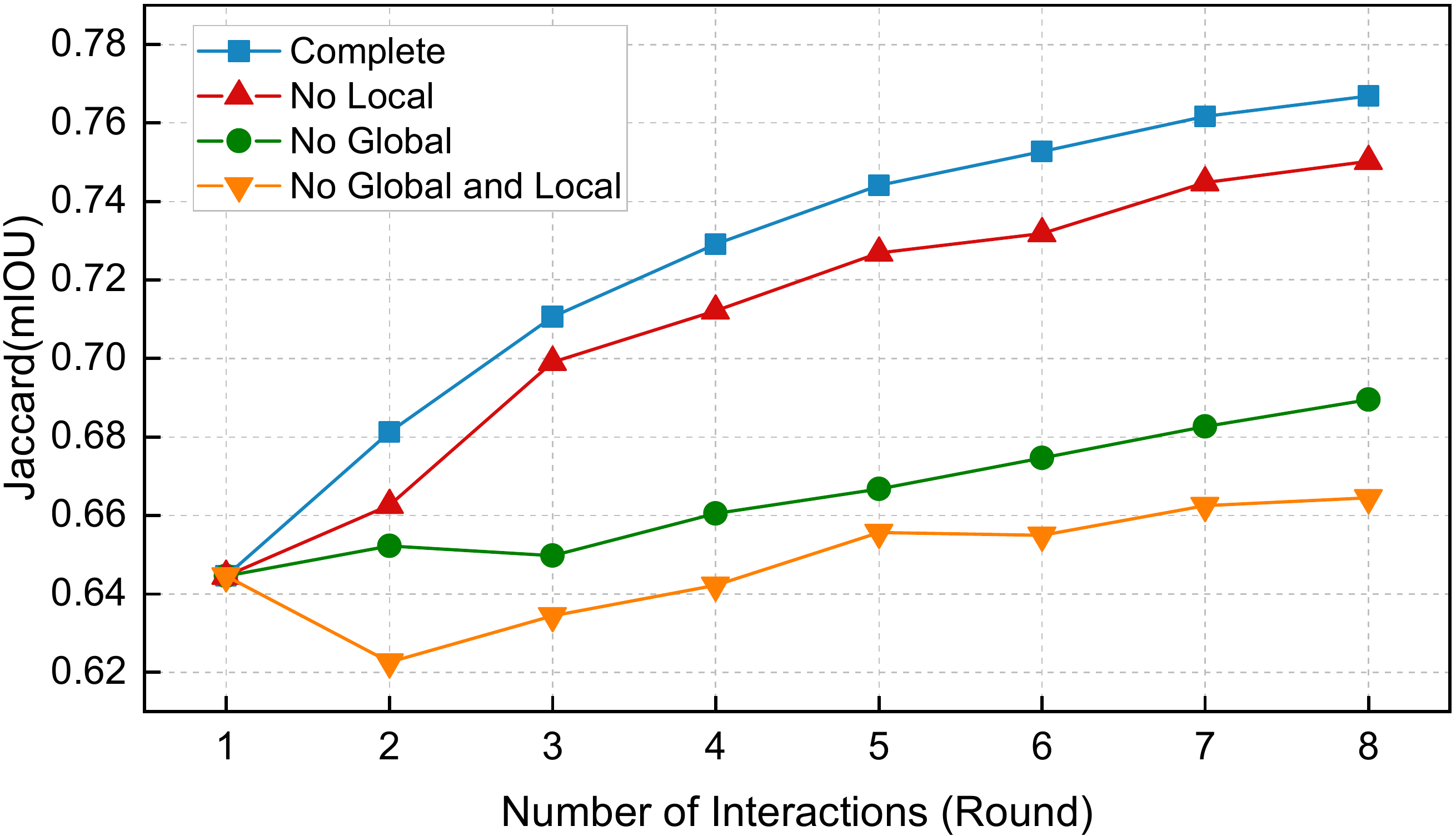}
\centering
\caption{ Ablation study on DAVIS 2017 validation set to show the effectiveness of our proposed global and local memories.}
\vspace{-1.5em}
\label{fig:ab1}
\end{figure}

\begin{figure}[t]
\includegraphics[width=0.85\linewidth]{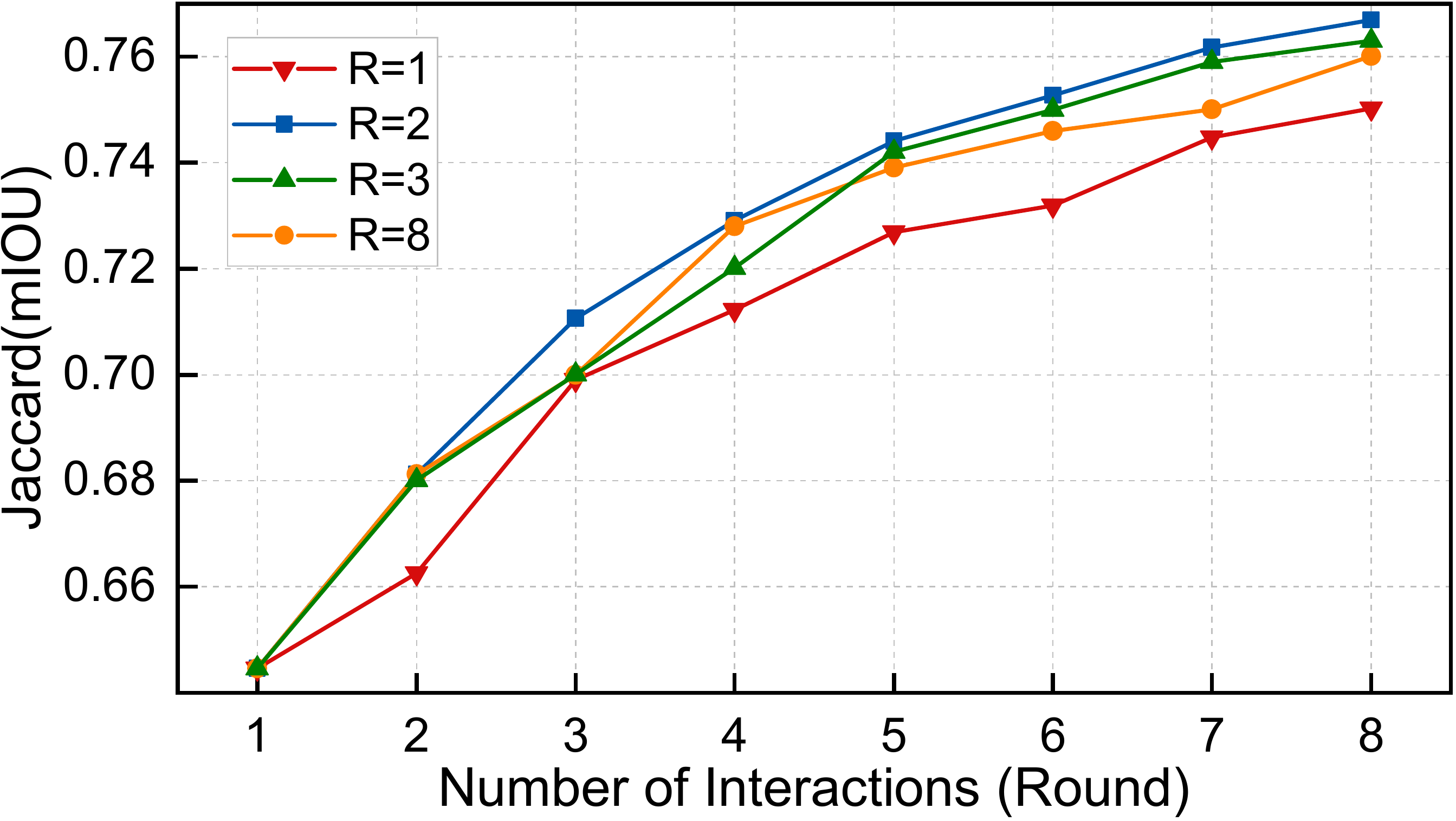}
\centering
\caption{The impact of $R$ in the local map memory. $R$ denotes that local maps in past $R$ rounds in the memory are used.}
\vspace{-1.5em}
\label{fig:ab2}
\end{figure}

\textbf{The Effectiveness of the Memory Mechanism. }
We conduct ablation studies using the DAVIS 2017 validation dataset to validate the effectiveness of our proposed memory mechanism. Fig.~\ref{fig:ab1} and Fig.~\ref{fig:ab2} show the Jaccard score of ablation models with growing number of interactions. In  Fig.~\ref{fig:ab1}, we compare our method with and without global and local memories. \textbf{No Global} indicates we use the model without the global memory, which means we only use the global map calculated in the first round and do not aggregate it in the following rounds. \textbf{No Local} indicates that we only use the local map calculated in the current round and do not access local maps from previous rounds. \textbf{No Global and Local} is a model without using both the global map memory and local map memory. We can find that both the global map memory and the local map memory take effects in the iVOS and greatly improves the performance since utilizing all scribble information of previous rounds. 

As described in Section 3, for the memory of the local map, there is a trade-off between choosing the nearest frame and the closest round. In practice, the segmentation mask far away from the annotated frame achieves worse results due to the error accumulation during propagation, so we choose the local map in which round it is nearest to the annotated frame. However, with the interactive round grows up, the accuracy of segmentation becomes better and better.
Therefore, we use the nearest map to the annotated frame in the past $R$ rounds. $R=1$ means we only use local maps of the current round while $R=8$ means we use the nearest map in all previous rounds. Fig.~\ref{fig:ab2} shows that when $R>1$, the segmentation accuracy will improve, indicating the effectiveness of the local map memory. When $R=2$, our method achieves the best performance, and we choose $R=2$ for our final model.

\textbf{The Effectiveness of the Augmented Map. }
 The augmented map of the interactive frame is stored in the global memory in the current interaction round, which will help this frame be correct segmented in the following interaction rounds. Therefore, without the augmented map, the valuable interactive information of this frame will be lost during the propagation in the following interaction rounds.
Besides, since our MA-Net also takes local matching into account, the improvements of all the interactive frames will further implicitly bring additional benefits to their subsequent non-interactive frames during the propagation. To be specific, the AUC score will drop from $0.749$ to $0.744$ if the augment map is removed from the global memory.

\textbf{The Impact of the Local Window Size. }
In addition, we also study the impact of the local window size $k$, as shown in Table.~\ref{table:localk}. When $k$ is smaller, the local map computation is more efficient. However, a small $k$ will affect the accuracy of our model. In practice, we choose $k=12$ in this paper.


\section{Conclusion}
Video object segmentation (VOS) is a fundamental task in computer vision. In this paper, we propose a user-friendly framework to generate accurate segmentation masks of a video with a few user annotations. Our MA-Net integrates the interaction and propagation operations into a unified pixel embedding learning framework, which promotes the efficiency of the round-based interactive VOS. More importantly, we propose a novel memory aggregation mechanism to record and aggregate the information of the user interactions and predictions of previous interaction rounds, which improves the segmentation accuracy greatly. 


\paragraph{Acknowledgments}

This work is in part supported by ARC DP200100938 and ARC DECRA DE190101315.

{\small
\bibliographystyle{ieee_fullname}
\bibliography{egbib}

\begin{thebibliography}{10}\itemsep=-1pt

\bibitem{bai2009video}
Xue Bai, Jue Wang, David Simons, and Guillermo Sapiro.
\newblock Video snapcut: robust video object cutout using localized
  classifiers.
\newblock {\em ACM Transactions on Graphics (ToG)}, 28(3):70, 2009.

\bibitem{bao2018cnn}
Linchao Bao, Baoyuan Wu, and Wei Liu.
\newblock Cnn in mrf: Video object segmentation via inference in a cnn-based
  higher-order spatio-temporal mrf.
\newblock In {\em CVPR}, pages 5977--5986, 2018.

\bibitem{benard2017interactive}
Arnaud Benard and Michael Gygli.
\newblock Interactive video object segmentation in the wild.
\newblock {\em arXiv preprint arXiv:1801.00269}, 2017.

\bibitem{bratt2012rotoscoping}
Benjamin Bratt.
\newblock {\em Rotoscoping}.
\newblock Routledge, 2012.

\bibitem{caelles2017one}
Sergi Caelles, Kevis-Kokitsi Maninis, Jordi Pont-Tuset, Laura Leal-Taix{\'e},
  Daniel Cremers, and Luc Van~Gool.
\newblock One-shot video object segmentation.
\newblock In {\em CVPR}, pages 221--230, 2017.

\bibitem{caelles20182018}
Sergi Caelles, Alberto Montes, Kevis-Kokitsi Maninis, Yuhua Chen, Luc Van~Gool,
  Federico Perazzi, and Jordi Pont-Tuset.
\newblock The 2018 davis challenge on video object segmentation.
\newblock {\em arXiv preprint arXiv:1803.00557}, 2018.

\bibitem{chen2018encoder}
Liang-Chieh Chen, Yukun Zhu, George Papandreou, Florian Schroff, and Hartwig
  Adam.
\newblock Encoder-decoder with atrous separable convolution for semantic image
  segmentation.
\newblock In {\em ECCV}, pages 801--818, 2018.

\bibitem{chen2018blazingly}
Yuhua Chen, Jordi Pont-Tuset, Alberto Montes, and Luc Van~Gool.
\newblock Blazingly fast video object segmentation with pixel-wise metric
  learning.
\newblock In {\em CVPR}, pages 1189--1198, 2018.

\bibitem{deng2009imagenet}
Jia Deng, Wei Dong, Richard Socher, Li-Jia Li, Kai Li, and Li Fei-Fei.
\newblock Imagenet: A large-scale hierarchical image database.
\newblock In {\em CVPR}, pages 248--255. Ieee, 2009.

\bibitem{fathi2017semantic}
Alireza Fathi, Zbigniew Wojna, Vivek Rathod, Peng Wang, Hyun~Oh Song, Sergio
  Guadarrama, and Kevin~P Murphy.
\newblock Semantic instance segmentation via deep metric learning.
\newblock {\em arXiv preprint arXiv:1703.10277}, 2017.

\bibitem{he2016deep}
Kaiming He, Xiangyu Zhang, Shaoqing Ren, and Jian Sun.
\newblock Deep residual learning for image recognition.
\newblock In {\em CVPR}, pages 770--778, 2016.

\bibitem{DAVIS2019-Interactive-2nd}
Yuk Heo, Yeong~Jun Koh, and Chang-Su Kim.
\newblock Interactive video object segmentation using sparse-to-dense networks.
\newblock {\em CVPR Workshops}, 2019.

\bibitem{hu2018videomatch}
Yuan-Ting Hu, Jia-Bin Huang, and Alexander~G Schwing.
\newblock Videomatch: Matching based video object segmentation.
\newblock In {\em ECCV}, pages 54--70, 2018.

\bibitem{krahenbuhl2013parameter}
Philipp Kr{\"a}henb{\"u}hl and Vladlen Koltun.
\newblock Parameter learning and convergent inference for dense random fields.
\newblock In {\em International Conference on Machine Learning}, pages
  513--521, 2013.

\bibitem{li2016roto++}
Wenbin Li, Fabio Viola, Jonathan Starck, Gabriel~J Brostow, and Neill~DF
  Campbell.
\newblock Roto++: Accelerating professional rotoscoping using shape manifolds.
\newblock {\em ACM Transactions on Graphics (TOG)}, 35(4):62, 2016.

\bibitem{li2018video}
Xiaoxiao Li and Chen Change~Loy.
\newblock Video object segmentation with joint re-identification and
  attention-aware mask propagation.
\newblock In {\em ECCV}, pages 90--105, 2018.

\bibitem{lin2014microsoft}
Tsung-Yi Lin, Michael Maire, Serge Belongie, James Hays, Pietro Perona, Deva
  Ramanan, Piotr Doll{\'a}r, and C~Lawrence Zitnick.
\newblock Microsoft coco: Common objects in context.
\newblock In {\em ECCV}, pages 740--755. Springer, 2014.

\bibitem{luiten2018premvos}
Jonathon Luiten, Paul Voigtlaender, and Bastian Leibe.
\newblock Premvos: Proposal-generation, refinement and merging for video object
  segmentation.
\newblock In {\em ACCV}, pages 565--580. Springer, 2018.

\bibitem{maninis2018video}
K-K Maninis, Sergi Caelles, Yuhua Chen, Jordi Pont-Tuset, Laura Leal-Taix{\'e},
  Daniel Cremers, and Luc Van~Gool.
\newblock Video object segmentation without temporal information.
\newblock {\em TPAMI}, 41(6):1515--1530, 2018.

\bibitem{DAVIS2018-Interactive-2nd}
Mohammad Najafi, Viveka Kulharia, T Ajanthan, and PH Torr.
\newblock Similarity learning for dense label transfer.
\newblock {\em CVPR Workshops}, 2018.

\bibitem{oh2019fast}
Seoung~Wug Oh, Joon-Young Lee, Ning Xu, and Seon~Joo Kim.
\newblock Fast user-guided video object segmentation by
  interaction-and-propagation networks.
\newblock In {\em CVPR}, pages 5247--5256, 2019.

\bibitem{oh2019video}
Seoung~Wug Oh, Joon-Young Lee, Ning Xu, and Seon~Joo Kim.
\newblock Video object segmentation using space-time memory networks.
\newblock In {\em ICCV}, 2019.

\bibitem{pohlen2017full}
Tobias Pohlen, Alexander Hermans, Markus Mathias, and Bastian Leibe.
\newblock Full-resolution residual networks for semantic segmentation in street
  scenes.
\newblock In {\em CVPR}, pages 4151--4160, 2017.

\bibitem{pont20172017}
Jordi Pont-Tuset, Federico Perazzi, Sergi Caelles, Pablo Arbel{\'a}ez, Alex
  Sorkine-Hornung, and Luc Van~Gool.
\newblock The 2017 davis challenge on video object segmentation.
\newblock {\em arXiv preprint arXiv:1704.00675}, 2017.

\bibitem{price2009livecut}
Brian~L Price, Bryan~S Morse, and Scott Cohen.
\newblock Livecut: Learning-based interactive video segmentation by evaluation
  of multiple propagated cues.
\newblock In {\em ICCV}, pages 779--786. IEEE, 2009.

\bibitem{ventura2019rvos}
Carles Ventura, Miriam Bellver, Andreu Girbau, Amaia Salvador, Ferran Marques,
  and Xavier Giro-i Nieto.
\newblock Rvos: End-to-end recurrent network for video object segmentation.
\newblock In {\em CVPR}, pages 5277--5286, 2019.

\bibitem{voigtlaender2019feelvos}
Paul Voigtlaender, Yuning Chai, Florian Schroff, Hartwig Adam, Bastian Leibe,
  and Liang-Chieh Chen.
\newblock Feelvos: Fast end-to-end embedding learning for video object
  segmentation.
\newblock In {\em CVPR}, pages 9481--9490, 2019.

\bibitem{voigtlaender2017online}
Paul Voigtlaender and Bastian Leibe.
\newblock Online adaptation of convolutional neural networks for the 2017 davis
  challenge on video object segmentation.
\newblock In {\em CVPR Workshops}, volume~5, 2017.

\bibitem{wang2005interactive}
Jue Wang, Pravin Bhat, R~Alex Colburn, Maneesh Agrawala, and Michael~F Cohen.
\newblock Interactive video cutout.
\newblock In {\em ACM Transactions on Graphics (ToG)}, volume~24, pages
  585--594. ACM, 2005.

\bibitem{wang2019learning}
Wenguan Wang, Hongmei Song, Shuyang Zhao, Jianbing Shen, Sanyuan Zhao,
  Steven~CH Hoi, and Haibin Ling.
\newblock Learning unsupervised video object segmentation through visual
  attention.
\newblock In {\em CVPR}, pages 3064--3074, 2019.

\bibitem{wang2019ranet}
Ziqin Wang, Jun Xu, Li Liu, Fan Zhu, and Ling Shao.
\newblock Ranet: Ranking attention network for fast video object segmentation.
\newblock In {\em ICCV}, 2019.

\bibitem{wug2018fast}
Seoung Wug~Oh, Joon-Young Lee, Kalyan Sunkavalli, and Seon Joo~Kim.
\newblock Fast video object segmentation by reference-guided mask propagation.
\newblock In {\em CVPR}, pages 7376--7385, 2018.

\bibitem{xu2018youtube}
Ning Xu, Linjie Yang, Yuchen Fan, Jianchao Yang, Dingcheng Yue, Yuchen Liang,
  Brian Price, Scott Cohen, and Thomas Huang.
\newblock Youtube-vos: Sequence-to-sequence video object segmentation.
\newblock In {\em ECCV}, pages 585--601, 2018.

\bibitem{xu2019mhp}
Shuangjie Xu, Daizong Liu, Linchao Bao, Wei Liu, and Pan Zhou.
\newblock Mhp-vos: Multiple hypotheses propagation for video object
  segmentation.
\newblock In {\em CVPR}, pages 314--323, 2019.

\bibitem{yang2018efficient}
Linjie Yang, Yanran Wang, Xuehan Xiong, Jianchao Yang, and Aggelos~K
  Katsaggelos.
\newblock Efficient video object segmentation via network modulation.
\newblock In {\em CVPR}, pages 6499--6507, 2018.

\bibitem{yang2020collaborative}
Zongxin Yang, Yunchao Wei, and Yi Yang.
\newblock Collaborative video object segmentation by foreground-background
  integration.
\newblock {\em arXiv preprint arXiv:2003.08333}, 2020.

\end{thebibliography}
}

\end{document}